\documentclass[10pt, a4paper]{article}
\usepackage{lrec}
\usepackage{graphicx}
\usepackage{tabularx}
\usepackage{soul}
\usepackage[utf8]{inputenc}
\usepackage{epstopdf}
\usepackage{epsf}
\usepackage{graphicx, tabularx}
\usepackage{url}
\usepackage{mathtools}
\usepackage{multirow, booktabs}
\usepackage{color}

\usepackage{flushend}
\newcommand{\corpus}{GeBioCorpus}
\newcommand{\toolkit}{GeBioToolkit}

\newcommand{\Ni}{({\em i})~}
\newcommand{\Nii}{({\em ii})~}
\newcommand{\Niii}{({\em iii})~}

\usepackage{hyperref}
\usepackage{xstring}
\usepackage{comment}

\title{\toolkit: Automatic Extraction of Gender-Balanced Multilingual Corpus of Wikipedia Biographies\\ \vspace*{.5\baselineskip}}

\name{Marta R. Costa-juss\`a, Pau Li Lin, Cristina Espa\~na-Bonet$^*$}

\address{TALP Research Center, Universitat Polit\`ecnica de Catalunya, Barcelona \\
$^*$ DFKI GmBH and Saarland University, Saarbr\"ucken \\
        \tt{ marta.ruiz@upc.edu, lilin.pau@gmail.com, cristinae@dfki.de}\\}

\abstract{
We introduce  \toolkit, a tool for extracting multilingual parallel corpora at sentence level, with document and gender information from Wikipedia biographies. Despite the gender inequalities present in Wikipedia, the toolkit has been designed to extract corpus balanced in gender. 
While our toolkit is customizable to any number of languages (and different domains), in this work we present a corpus of 2,000 sentences in English, Spanish and Catalan, which has been post-edited by native speakers to become a high-quality dataset for machine translation evaluation. 
While \corpus\ aims at being one of the first non-synthetic gender-balanced test datasets, \toolkit\ aims at paving the path to standardize procedures to produce gender-balanced datasets.
\\ \newline }
\begin{document}

\maketitleabstract

\section{Introduction}

Gender biases are present in many natural language processing applications \cite{costajussa:2019}. This comes as an undesired characteristic of deep learning architectures where their outputs seem to reflect demographic asymmetries~\cite{pratesEtAl:2018}. This is of course not due to the architecture itself but to the data used to train a system. 
Recent research is being devoted to correct the asymmetries mainly by data augmentation techniques in fields such as coreference resolution \cite{rudingerEtAl:2018,zhaoEtAl:2018,websterEtAl:2018} or abusive language detection \cite{parkEtAl:2018}. Test sets have been created in those cases, but we are not aware of any test set available for machine translation (MT).

From another side, machine translation either neural, statistical or rule-based, usually operates in a sentence-by-sentence basis. However, when translating consistently a document, surrounding sentences may have valuable information. The translation of pronouns, verb tenses and even content words might depend on other fragments within a document. This affects also the translation of the gender markers, specially when translating from a language without these marks (e.g. English) into a language with them (e.g. Catalan).
Test sets at sentence level are not useful to evaluate these phenomena. But in the time of claims of human parity in MT, all these phenomena are crucial, and document-level evaluation is needed in order to discriminate among systems \cite{lubliEtAl:2018}. 

Beyond these gender-balanced and document-level needs, the rise of multilingual neural machine translation and the lack of multi-way parallel corpus that can evaluate its abilities (e.g. zero-shot), motivates the creation of new multilingual corpora. In turn, this motivation prompts the development of software to automatically create such corpora.

In order to create a \emph{\toolkit\ }that is able to systematically extract multilingual parallel corpus at sentence level and with document-level information, we rely on Wikipedia, a free online multilingual encyclopedia written by volunteers. The toolkit is customizable for languages and gender-balance. We take advantage of Wikipedia multilinguality to extract a corpus of biographies, being each biography a document available in all the selected languages. However, given the bias in the representation of males and females also in Wikipedia \cite{bammanSmith:2014} and, to deliver a balanced set, we need to identify and select a subset of documents just after the toolkit performs the extraction of parallel sentences so it can assure parity. Note that our document-level extraction is consistent (all sentences in a document belong to the same personality) but the documents do not keep coherence anymore, since we are removing some sentences within that document. In our experiments in English--Spanish--Catalan, \toolkit\ is able to extract parallel sentences with 87.5\% accuracy according to a human evaluation.


Besides providing the tool, we also manually post-edited a small subset of the English (en), Spanish (es) and Catalan (ca) outputs to provide a test dataset which is ready-to-use for machine translation applications; to our knowledge the first 
gender-balanced dataset extracted from real texts in the area: \emph{\corpus}. 
With these three languages we aim to cover two linguistic families (Germanic and Romance)
which differ in morphology and specifically in gender inflection. The choice in languages is intended to allow the evaluation of machine translation outputs in three different settings: distant morphologies for a high-resourced language pair (English--Spanish) and a low-resourced pair (English--Catalan), and closely related languages (Spanish--Catalan).
%
The latter responds the challenge recently initiated in the WMT International Evaluation%
\footnote{\url{http://www.statmt.org/wmt19/similar.html}}. 
We used native/fluent speakers to provide post-editions on the final multilingual set in the three languages.

The rest of the paper is organized as follows. Section~\ref{s:related} describes some available multilingual datasets used for machine translation evaluation, related work on parallel sentence extraction from Wikipedia and a brief mention to general research lines on gender bias in NLP. 
Section~\ref{sec:toolkit} describes the architecture of the tool and Section~\ref{s:corpus} the methodology, evaluation and characteristics of the extracted corpora. Finally, Section~\ref{s:conclusions} summarizes the work and points at several future extensions of \toolkit.

\section{Related Work}
\label{s:related}

There are several multilingual parallel datasets available to evaluate MT outputs. The corpora covering more languages are JRC-Acquis (Acquis Communautaire) and the TED talks corpus. 
Arab-Acquis \cite{arabAcquis:2017} is a multilingual dataset for 22 European languages plus Arabic with more than 12,000 sentences coming from JRC-Acquis, that is, a collection of legislative texts of the European Union. In order to make the test set equal in all the languages, only sentences that are parallel simultaneously in the 22 languages were extracted \cite{koehnEtAl:2009} and, therefore, the document structure of the data is lost.  

The Web Inventory of Transcribed and Translated Talks, WIT$^{3(}$%
\footnote{\url{https://wit3.fbk.eu/}}$^{)}$, includes English subtitles from TED talks and their translations currently in 109 languages. Parallel corpora are extracted for several pairs \cite{cettoloEtAl:2012} and test sets are annually prepared for the International Workshop on Spoken Language Translation (IWSLT) evaluation campaigns. Test sets exist for all the pairs among German, English, Italian, Dutch and Romanian; and from English to Arabic, Basque, Chinese, Czech, Dutch, Farsi, French, German, Hebrew, Italian, Japanese, Korean, Polish, Portuguese-Brazil, Romanian, Russian, Slovak, Slovenian, Spanish, Thai, Turkish and Vietnamese. In this case the whole talks are aligned at sentence level, so, the document structure is kept but the set is not equivalent in all the languages.

Similarly, the news translation task at the annual workshops and conferences on statistical machine translation (WMT) distributes collections of parallel news to their participants. Test sets have been created over the years for pairs including English into Chinese, Czech, Estonian, Finnish, French, German, Hindi, Hungarian, Kazakh, Latvian, Spanish, Romanian, Russian and Turkish. Again, the document structure is kept in the dataset but, in general, the set is not equivalent in all the languages.

Wikipedia is widely used in natural language processing and it is an excellent resource for multilingual tasks, parallel sentence extraction being among them. However, we do not know of any multilingual corpus of biographies extracted from the resource. On the monolingual side, one can find  WikiBiography%
\footnote{\url{https://www.h-its.org/en/research/nlp/wikibiography-corpus/}},  a corpus of 1,200 biographies in German with automatic annotations of PoS, lemmas, syntactic dependencies, anaphora, discourse connectives, classified named entities and temporal expressions. 
The authors in \newcite{bammanSmith:2014} also extract 927,403 biographies in this case from the English Wikipedia. The set is  pre-processed in order to learn event classes in biographies.

Regarding the automatic extraction of parallel corpora, Wikipedia has been traditionally used as a resource.
In \newcite{Adafre:06}, the authors extract parallel sentences based on the available metadata in Wikipedia texts.
Both \newcite{Yasuda:08} and \newcite{Plamada:12} extracted parallel sentences by translating the articles into a common language and consider those sentences with a high translation quality to be parallel. The ACCURAT project~\cite{Stefanescuetal:12,Skadinaetal:12} also devoted efforts in parallel sentence mining in Wikipedia. Later, \newcite{BarronEtAl:2015} used the combination of cross-lingual similarity measures to extract domain specific parallel sentences.
The most recent initiative is the so-called LASER~\cite{artetxeSchwenk:2019}, which relies on vector representations of sentences to extract similar pairs. This toolkit has been used to extract the WikiMatrix corpus \cite{DBLP:journals/corr/abs-1907-05791} which contains 135 million parallel sentences for 1,620 different language pairs in 85 different languages. 


As far as we are concerned, there is no gender-balanced dataset for machine translation, except for the artificial gold standard created for English--Spanish \cite{font:2019}.
However, there are a notable number of works towards doing research in the area: from balancing data sets in monolingual tasks \cite{websterEtAl:2018,rudingerEtAl:2018} to evaluating gender bias in several tasks \cite{basta-etal-2019-evaluating,stanovsky-etal-2019-evaluating}.

\begin{figure*}[h!]
\begin{center}
  \includegraphics[width=0.75\textwidth]{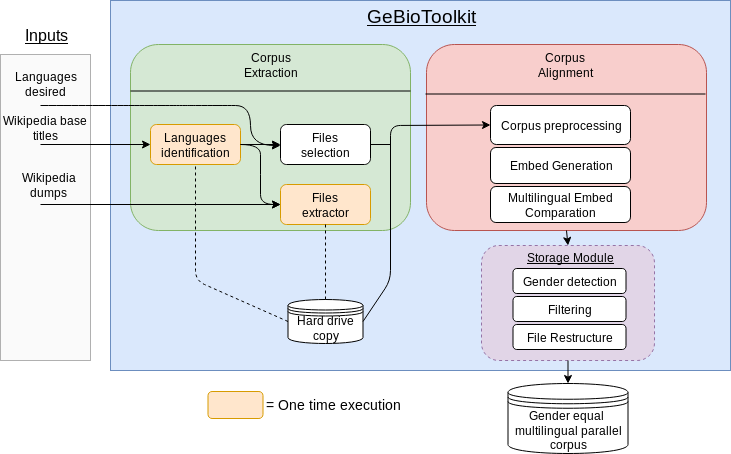}
  \caption{\toolkit\ architecture}\label{fig:gebiocorpus}
 \end{center}
\end{figure*}

\section{\toolkit}
\label{sec:toolkit}

\subsection{Base Architecture}
\label{ss:method}

To extract the corpus previously introduced, we develop \toolkit.
The tool retrieves (multi-)parallel sentences for any Wikipedia category 
and for any arbitrary number of languages at the same time. In its default configuration, the toolkit retrieves gender-balanced corpora. \toolkit\ is composed by three blocks.
The first block, corpus extractor, provides a layer to transform, collect and select entries in the desired languages. The second block, corpus alignment, finds the parallel sentences within a text and provides a quality check of the parallel sentences given a few restrictions. The third block, gender classifier,
includes a filtering module which classifies the gender of the entry and outputs the final parallel corpus. The gender detection functionality can be activated or deactivated at will, allowing the tool to be used in a more general context, where gender-balanced data is either non-relevant or not needed. Figure~\ref{fig:gebiocorpus} depicts its architecture.

The tool requires three inputs: \Ni a list of the desired languages, \Nii the dump files for the languages%
\footnote{Dumps can be downloaded from\\ \hspace*{1.7em} \url{https://dumps.wikimedia.org/}} and \Niii a list of the articles' titles belonging to the category to extract (currently in English).
For our purpose of gathering a corpus of biographies, we retrieve the list of articles that belong to the "living people" category by using the PetScan tool%
\footnote{\url{https://petscan.wmflabs.org/}}.


The \textbf{corpus extractor} module starts by looking for the equivalent articles to those input in the other languages via the Wikipedia interlanguage links.
The module also retrieves the multilingual titles
providing a dictionary of \{english\_entry\_title: [language, title]\} that is used on the file extraction and file selection modules. File extraction retrieves then the selected entries in the previous dictionary from the Wikipedia dump. 
\toolkit\ uses
a modified version of the \textit{wikiextractor}\footnote{\url{https://github.com/attardi/wikiextractor}} software
to retrieve and store the different Wikipedia entries from each language. 
Finally, file selection generates a dictionary similar to the one obtained before,
but it only stores the entries for which the files were successfully retrieved.

The \textbf{corpus alignment} module makes use of the text and dictionaries retrieved in the previous step and the LASER toolkit%
\footnote{\url{https://github.com/facebookresearch/LASER}}. LASER  (Language-Agnostic SEntence Representations) allows to obtain sentence embeddings through a multilingual sentence encoder \cite{DBLP:journals/corr/abs-1907-05791}. 
Translations can be found then as close pairs (tuples) in the multilingual semantic space.
In order to perform the parallel sentence extraction, we follow the margin-based criterion introduced in \newcite{artetxeSchwenkMargin}.
%
The margin between two candidate sentences $x$ and $y$ is defined as the ratio between the cosine distance between the two embedded sentences, and the average cosine similarity of its nearest neighbors in both directions:

$margin(x,y) = \frac{\cos{(x,y)}}{ \sum\limits_{\substack{z\in NN_k(x)}} \frac{\cos(x,z)}{2k} + \sum\limits_{\substack{z\in NN_k(y)}} \frac{\cos(y,z)}{2k}},$

\noindent
where $NN_k(x)$ denotes the $k$ unique nearest neighbors of $x$ in the other language, $NN_k(y)$ the same for $y$. 

To extract parallel sentences on more than two languages, let us say $i$ languages, we use a greedy approach with a pivot language $L_1$. We detect all the parallel sentences in the pairs $L_1$--$L_i$ and then extract the intersection of sentences between the language pairs.




\begin{table*}[h]
\begin{center}
\scriptsize
\scalebox{0.9}{
\begin{tabular}{ |l|l } 
\hline
~\\
\textless doc docid="Aurelia Arkotxa " wpid="51690640" language="en" topic="C6" gender="Female" \textgreater \\
\textless title\textgreater Aurelia Arkotxa \textless /title\textgreater \\
\textless seg id="1"\textgreater She teaches classics at the University of Bayonne; she was co-founder of the literary magazine and a new newspaper.\textless \textbackslash seg\textgreater \\
\textless /doc\textgreater \\
\textless doc docid="Catriona Gray " wpid="51838666" language="en" topic="C2" gender="Female"\textgreater\\
\textless title\textgreater Catriona Gray \textless /title\textgreater\\
\textless seg id="1"\textgreater In addition, she obtained a certificate in outdoor recreation and a black belt in Choi Kwang-Do martial arts.\textless \textbackslash seg \textgreater\\
\textless seg id="2\textgreater Catriona Elisa Magnayon Gray (born 6 January 1994) is a Filipino-Australian model, singer, and beauty pageant titleholder who was crowned Miss Universe 2018.\textless \textbackslash seg\textgreater\\
\textless seg id="3"\textgreater Gray was born in Cairns, Queensland, to a Scottish-born father, Ian Gray, from Fraserburgh, and a Filipina mother, Normita Ragas Magnayon, from Albay.\textless \textbackslash seg \textgreater\\
\textless /doc\textgreater \\
\tiny{~}\\
\hline
~\\
\textless doc docid="Aurelia Arkotxa" wpid="7789214" language="es" topic="C6" gender="Female" \textgreater \\
\textless title\textgreater Aurelia Arkotxa\textless /title \textgreater \\
\textless seg id="1"\textgreater Enseña cultura clásica en la facultad de Bayona; fue cofundadora de una revista literaria y de un diario.\textless \textbackslash seg\textgreater \\ 
\textless/doc\textgreater \\
\textless doc docid="Catriona Gray" wpid="8411924" language="es" topic="C2" gender="Female" \textgreater \\
\textless title\textgreater Catriona Gray \textless /title \textgreater \\
\textless seg id="1"\textgreater Además, obtuvo un Certificado en Recreación al Aire Libre y un cinturón negro en Artes Marciales de Choi Kwang-Do.\textless \textbackslash seg\textgreater \\
\textless seg id="2"\textgreater Catriona Elisa Magnayon Gray (6 de enero de 1994) es una modelo y reina de belleza australiana-filipina, ganadora de Miss Universo 2018 representando a Filipinas.\textless \textbackslash seg\textgreater \\
\textless seg id="3"\textgreater Gray nació en Cairns, Queensland de un padre australiano nacido en Escocia, Ian Gray, de y una madre filipina, Normita Ragas Magnayon, de Albay.\textless \textbackslash seg\textgreater \\
\textless /doc\textgreater \\
\tiny{~}\\
\hline

\end{tabular}
}
\caption{Example of two documents extracted in English (top) and the parallel counterparts in Spanish (bottom) from GeBioCorpus-v2.}
\label{tab:example2}
\end{center}
\end{table*}

\subsection{Gender Detection Module}
\label{ss:gender}

The previous two blocks implement a general approach to parallel sentence extraction following a similar methodology as that used to extract the WikiMatrix corpus. But for our purpose, we need to specifically deal with gender bias.
Wikipedia is known to have a gender bias in its content. Depending on the language and year of the study, the percentage of biographies of women with respect to the total of biographies ranges from a 13\% to a 23\% \cite{reagleRhue:2011,bammanSmith:2014,wagnerEtAl:2016}. 
And it is not only this, but also men and women are characterised differently \cite{Graells-GarridoEtAl:2015,wagnerEtAl:2016}, showing for instance more man in sports and woman in divorces \cite{reagleRhue:2011}. 

To allow extracting a gender-balanced dataset, we detect the gender automatically and filter files in order to have 50\% of articles for each gender. 
Following \newcite{reagleRhue:2011}, the gender of the article is extracted as that corresponding to the maximum number of gendered pronouns (i.e., he and she in English) mentioned in  the  article. According to their results,  this method of gender inference is overwhelmingly accurate,  in  a  random  test  set  of  500  articles,  it achieved 100\% precision with 97.6\% recall (12 articles had no pronominal mentions and so gender was not assigned).


\subsection{Cleaning Module}
\label{ss:postpro}


We analyse the accuracy of the extractions in Section~\ref{ss:eval}, but a simple visual inspection already shows a frequent source of noise in the extracted data that can be easily removed with a post-processing step. Some of the extracted sentence pairs include information in one language that is lacking in the other one(s).
For example, the sentence in Spanish \textit{"Mahrez se casó con su novia inglesa en 2015 y tuvieron una hija ese mismo año."} is aligned with the sentence in English \textit{"Mahrez married his English girlfriend Rita Johal in 2015."}, where the Spanish segment \textit{"y tuvieron una hija ese mismo año."} is not present in English. 
To avoid this, we filter out sentence pairs with large length ratio (filter samples in which one of the sentences is at least 20\% longer than the others). Such filtering can be generalised to distant language pairs by estimating language-dependent length factors \cite{pouliquen:2003}
or considering more elaborated corpus filtering techniques \cite{koehn-etal-2019-findings}. 

\subsection{File Restructure Module}
\label{ss:file}
Finally, the extracted parallel sentences are written as an xml file using a document-level mark-up. As said before, output documents do not preserve coherence any more, but document-level characteristics such as lexical consistency are kept and the mark-up allows use the information.

Table~\ref{tab:example2} shows an excerpt of the output file for English and Spanish. For each original biography, we keep the English title as document ID, the original Wikipedia ID, the extracted gender, and in our clean test set (see Section~\ref{s:corpus}) we also add the topic or category of each document. The \texttt{docid} field links
documents among languages.

\section{\corpus}
\label{s:corpus}

We use the \toolkit\ presented in the above section to extract a multilingual parallel corpus balanced in gender for English, Spanish and Catalan. The motivation for extracting this specific dataset is the variation in morphology in English (Germanic language) as compared to Spanish and Catalan (Romance languages). 
The variation in morphology is one of the most relevant challenges when solving biases in gender, see examples in \cite{pratesEtAl:2018,font:2019}.

\begin{figure*}[h]
\begin{center}
  \includegraphics[width=1\textwidth]{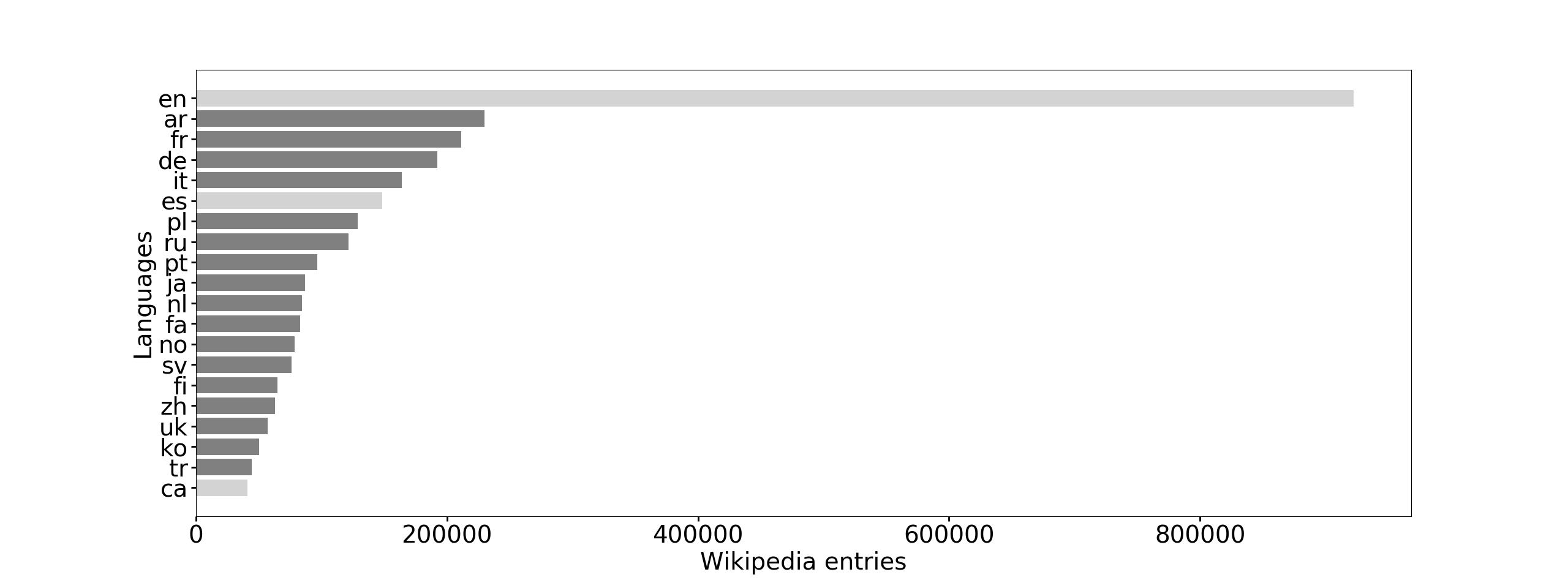}
  \caption{Number of Wikipedia entries under the "living people" category for the 20 Wikipedia editions with the most number of entries.}\label{fig:evollang}
 \end{center}
\end{figure*}
\begin{figure*}[h]
\begin{center}
  \includegraphics[width=0.95\textwidth]{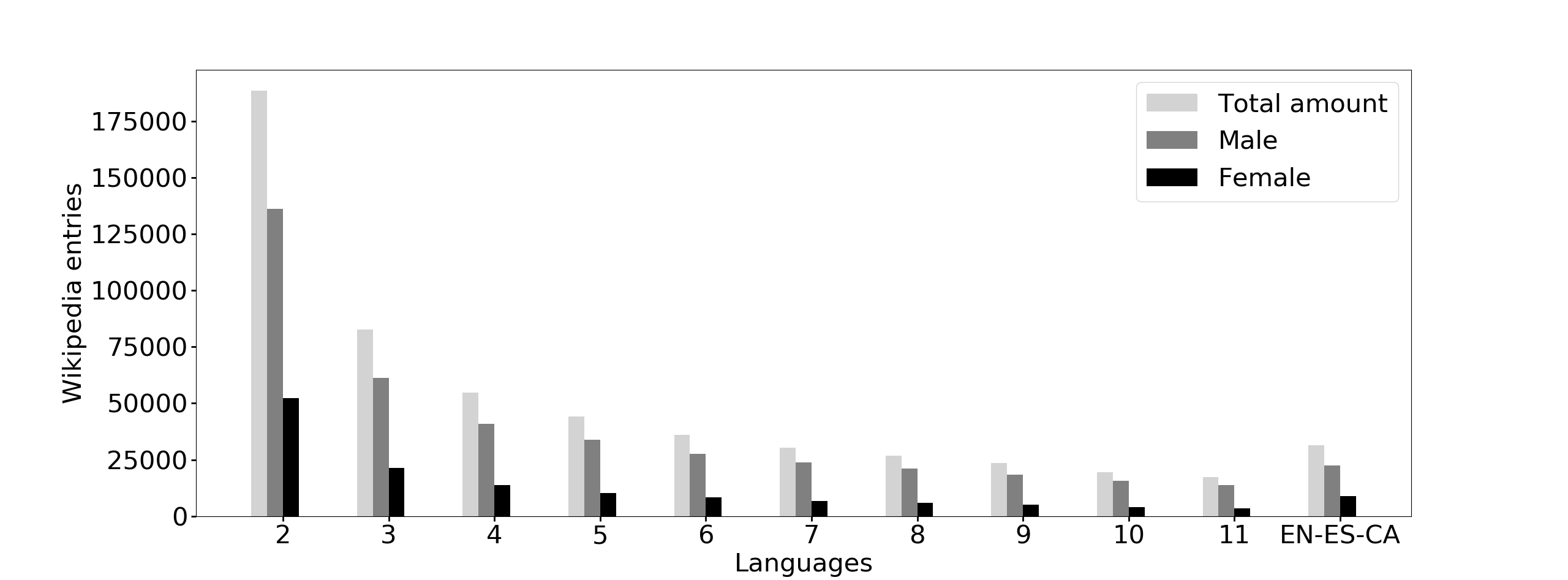}
  \caption{Number of documents by sets of languages under the Wikipedia category "living people" distributed by gender. See Section~\ref{ss:statistics} for the specific languages involved in the sets.}\label{fig:fileslang}
 \end{center}
\end{figure*}

\begin{table*}[h!]
\begin{center}
\scalebox{0.88}{
\footnotesize
\begin{tabular}{ l|l } 
 \hline
EN & In 2008, she was honored by the Angolan Ministry of Culture with a certificate of honor for her services to Angolan children's literature.\\ 
ES & En 2008, fue honrada por el Ministerio de Cultura de Angola con un certificado de honor por sus servicios a la literatura infantil angoleña.\\ 
CA & El 2008 va rebre el premi del Ministeri de Cultura angolès amb un certificat d'honor pels seus serveis a la literatura infantil angolesa.\\  \hline
EN & His poetry is characterized by a colloquial language and by his reflections regarding every day events or situations.\\
ES & Su poesía se caracteriza por un lenguaje coloquial y por la reflexión a partir de acontecimientos o situaciones cotidianas.\\
CA & La seva poesia es caracteritza per un llenguatge col·loquial i per la reflexió a partir d'esdeveniments o situacions quotidianes.\\ \hline
EN & Bridegroom was an actor and songwriter who hosted the TV series "The X Effect".\\
ES & Bridegroom era un actor y compositor quién participó de la serie de televisión "El Efecto X.\\
CA & Bridegroom era un actor i compositor que va participar de la sèrie de televisió "The X Effect."\\ \hline
EN & She was also recognised in 2015 by the British Council for her community work, and by the BBC as part of their "100 Women" series.\\
ES & También fue reconocida en 2015 por el British Council por su trabajo comunitario, y por la BBC como parte de su serie "100 Women (BBC)".\\
CA & També va ser reconeguda el 2015 pel British Council pel seu treball comunitari, i per la BBC com a part de la seva sèrie "100 Women (BBC)".\\ \hline
\end{tabular}}
\caption{Examples of parallel segments extracted for English, Spanish and Catalan from GeBioCorpus-v2.}\label{tab:example3}
\end{center}
\end{table*}

\begin{table*}[]
\begin{center}
\begin{tabular}{ |l|l|l|l|l|l|l| } 
 \hline
\emph{\corpus-v1} &  \multicolumn{2}{c|}{En} &  \multicolumn{2}{c|}{Es} &  \multicolumn{2}{c|}{Ca} \\ \hline
&F &M &F&M&F&M\\ \hline
Documents & 2287&  2741 &2287&  2741& 2287&  2741 \\ 
 \hline
Sentences &  {8000} & 8000 & 8000&8000&8000&8000 \\ 
 \hline
Average sent/doc & 3.5 &2.9&3.5&2.9&3.5& 2.9  \\ \hline
Words & 228.9k & 235.3k & 230.1k &236.0k& 240.9k& 245.7k\\ \hline
Average words/doc & 56.9 &51.1&56.3 &47.3&59.8&53.3\\ \hline
Vocabulary & 24.3k & 24.8k &27.1k & 27.6k & 27.6k &28.0k\\ \hline\hline
\emph{\corpus-v2} & \multicolumn{2}{c|}{En} &  \multicolumn{2}{c|}{Es} &  \multicolumn{2}{c|}{Ca} \\ \hline
&F &M &F&M&F&M\\ \hline
Documents & 257&339 & 257&339 & 257&339 \\ 
 \hline
Sentences &  1000 &1000&1000&1000&1000&1000 \\ 
 \hline
Average sent/doc &  3.9 &2.9 &3.9&2.9&3.9 & 2.9 \\ \hline
Words &28.7k &27.5k& 28.9k& 28.0k& 30.1k&29.4k\\ \hline
Average words/doc &  61.6&46.6&61.4 &46.4& 65.4& 49.7\\ \hline
Vocabulary &6.1k& 6.0k& 6.5k& 6.5k& 6.7k&6.6k\\ \hline
\end{tabular}
\caption{Statistics for the two corpora extracted by \toolkit\ without (\corpus-v1) and with (\corpus-v2) manual post-edition on the raw output. 
}\label{tab:statistics}
\end{center}
\end{table*}



\subsection{Data Statistics}
\label{ss:statistics}

We extract all the articles belonging to the category "living people" in the 40 largest Wikipedia editions to have a rough idea of the number of biographies available per language.
Figure \ref{fig:evollang} shows the amount of these articles for the 20 languages with the largest number of biographies.
The edition with the most available entries is the English one with 922,120 entries. 
The Spanish Wikipedia contains 148,445 entries (6th largest one) and the Catalan edition 40,983 (20th largest one). All three are highlighted in Figure \ref{fig:evollang}. Even if the Catalan Wikipedia is not as big as the English and Spanish ones, there is a noticeable amount of comparable articles  between Spanish and Catalan which translates into significant number of parallel sentences ---\newcite{DBLP:journals/corr/abs-1907-05791} extracted in WikiMatrix 3,377\,k sentences for en--es, 1,580\,k sentences for ca--es and 210\,k sentences for en--ca from the full editions.

\toolkit\ extracts multilingually aligned parallel sentences, so it is interesting to study also the union of languages. The more languages involved, the lesser number or comparable articles one will obtain.
Figure~\ref{fig:fileslang} shows the number of articles per sets of languages  and broken-down by gender. The total number of documents decays when starting with English and Arabic (set with only 2 languages) and one incrementally adds French, German, Italian, Spanish, Polish, Russian, Portuguese, Japanese and Dutch. With all languages (set with 11 languages), the number of comparable documents results in 17,285, out of which 13,676 are male bios and 3,609 are female bios. These numbers reconfirm the results by \cite{reagleRhue:2011,bammanSmith:2014,wagnerEtAl:2016} on the underrepresentation of women in Wikipedia.
Note that by doing the union of different languages and the intersection of its articles, the amount of documents decays quite abruptly, but the percentual difference between man and woman remains close to be constant. 

When we consider English, Spanish and Catalan, we retrieve 31,413 bibliographies.
\toolkit\ selects 81,405 sentences from them, 53,389 sentences are related to male bibliographies and 28,016 to female bibliographies. 
47.5\% of the male sentences  are removed to obtain a gender-balanced corpus. Then, \toolkit\ performs the length-based cleaning, which filters the corpus down to 16,679 sentences on male bibliographies and 10,730 sentences on female bibliographies. 

As output, we provide two versions of the corpus. \corpus-v1 contains 16,000 sentences as extracted from Wikipedia without any manual post-edition. \corpus-v2 contains 2,000 sentences with manual post-edition as explained in Section~\ref{ss:cats}. One set excludes the other in addition to excluding other sentences (1730 for female) that we plan to post-edit in the future. See statistics on number of documents, sentences, words and vocabulary broken-down per language and gender in Table \ref{tab:statistics}.
Table \ref{tab:example3} provides some examples of sentences from \corpus-v2.

\begin{figure}[h]
\centering
  \includegraphics[trim={1.5em 3.5em 0 0},clip,width=0.5\textwidth]{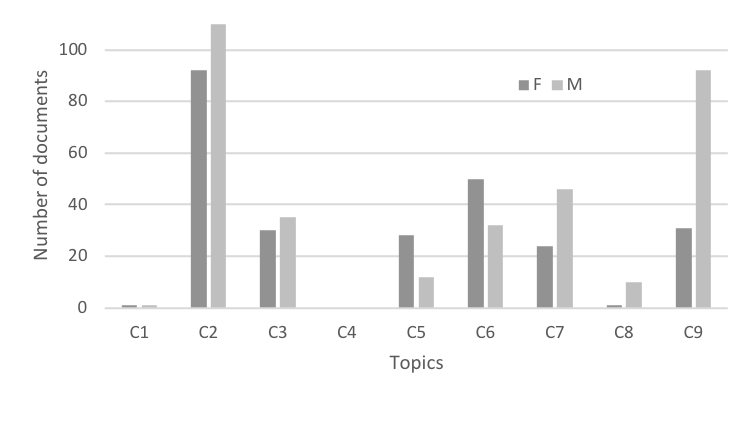}
  \caption{Distribution of topic/categories for the manually annotated test set. The distribution is further split by genre (articles of women (F) and men (M)). See Section~\ref{ss:cats} for the categories defined.} \label{fig:categories}
\end{figure}

\subsection{Manual Post-Editing and Categorization}
\label{ss:cats}

After the extraction of the complete corpus, and in order to provide a high-quality evaluation corpus, we ask native/fluent speakers to post-edit 2,000 sentences belonging to both female and male documents. Annotators were asked to either edit segments to make them parallel (if differences were small) or to delete segments that were too different. Also during the process, annotators were asked to provide a topic for the profession of each personality. Categories are based on the Wikipedia list of occupations\footnote{\url{https://en.wikipedia.org/wiki/Lists\_of\_occupations}}, which includes: (C1) Healthcare and medicine, (C2) Arts, (C3) Business, (C4) Industrial and manufacturing, (C5) Law enforcement, social movements and armed forces, (C6) Science, technology and education, (C7) Politics, (C8) Religion, and (C9) Sports. This corresponds to the field "topic" added in the header of each document as seen in Table \ref{tab:example2}. 
Figure \ref{fig:categories} shows the distribution of the 9 topics by gender. Note that the most unbalanced topics are C5 (Law enforcement, social movements and armed forces), C8 (Religion) and C9 (Sports).
We report other statistics for this test set in Table~\ref{tab:statistics}.

\subsection{Human Evaluation}
\label{ss:eval}

Finally, we perform a human evaluation on the quality of \corpus-v1, that is, the data as extracted by \toolkit\  without any post-edition. For this, we randomly select 50 sentences in the three languages (English, Spanish and Catalan) and present each tuple to 7 different native/fluent speakers. These annotators were asked to score a tuple with 1 if the three sentences convey the same meaning and 0 otherwise. 
On average, evaluators gave 87.5\% accuracy and when computing the majority vote among the evaluators accuracy reached 96\%. We computed Fleiss' kappa \cite{fleiss1971mns} as a measure for the interannotator agreement which resulted in 0.67, which is considered a substantial agreement
\footnote{\url{https://en.wikipedia.org/wiki/Fleiss\%27\_kappa}}. On average, our \toolkit\ plus filtering by length has 87.5\% accuracy. 

\section{Conclusions}
\label{s:conclusions}


This paper presents two main contributions. From one side, \toolkit, based on LASER, allows to automatically extract  multilingual parallel corpora at sentence-level which can be customized in number of languages and balanced in gender.
Document-level information is kept in the corpus and each document is tagged with the ID of the original Wikipedia article, the language and the gender of the person is referring to. From another side, we provide a multilingual corpus of biographies in English, Spanish and Catalan, \corpus. Two versions of this corpus are presented. The first version, \corpus-v1, contains 16k sentences which have been directly extracted using \toolkit. This version of the corpus is used to make an evaluation of the quality of the extractions produced by our tool. A manual evaluation shows that the accuracy in the multilingual parallel sentences is 87.5\%. The second version of this corpus, \corpus-v2, contains 2k sentences which have been post-edited by native speakers and categorized with type of profession. In this case, and in addition to the automatic tags of each document (ID, language and gender), each document is tagged with an occupation category. This labels allows to split the corpus in different subsets to evaluate domain-specific translations and gender accuracy for instance. 

As future improvements to the tool, we will remove the dependence on PetScan ---or any external tool to obtain lists of articles---, and extract all the necessary information from the input Wikipedia dump. We will also study the viability of using Wikidata information instead of the most frequent pronoun in a text in order to classify the gender of an article.


Both \toolkit\ and \corpus\ will be made available in github during the review process of this paper\footnote{\url{https://github.com/PLXIV/Gebiotoolkit}}.

\section*{Acknowledgments}
The authors want to thank Jordi Armengol, Magdalena Biesialska, Casimiro Carrino, Noe Casas, Guillem Cort\`es, Carlos Escolano, Gerard Gallego and Bardia Rafieian for their help in postediting the \corpus, Eduardo Lipperheide, Anna Li, Ayyoub Marroun, Laura Termes, Adrià Garcia, Alba Salomon and Clara Grenzner for the manual evaluation and Melissa Steiner, Ekaterina Lapshinova-Koltunski and Pauline Krielke for all the support and fruitful discussions. 

This work is supported in part by the Spanish Ministerio de Econom\'ia y Competitividad, the European Regional  Development  Fund  and  the  Agencia  Estatal  de  Investigaci\'on,  through  the  postdoctoral  senior grant Ram\'on y Cajal, the contract TEC2015-69266-P (MINECO/FEDER,EU) and the contract PCIN-2017-079 (AEI/MINECO). It is also partially funded by the German Federal Ministry of Education and Research under the funding code 01IW17001 (Deeplee). Responsibility for the content of this publication is with the authors.


\section{Bibliographical References}
\bibliographystyle{lrec}
\bibliography{lrec2020W-xample,paper}

\begin{thebibliography}{}

\bibitem[\protect\citename{Adafre and {de Rijke}}2006]{Adafre:06}
Adafre, S. and {de Rijke}, M.
\newblock (2006).
\newblock {Finding Similar Sentences across Multiple Languages in Wikipedia}.
\newblock In {\em {Proceedings of the 11th Conference of the European Chapter
  of the Association for Computational Linguistics}}, pages 62--69.

\bibitem[\protect\citename{Artetxe and Schwenk}2019a]{artetxeSchwenkMargin}
Artetxe, M. and Schwenk, H.
\newblock (2019a).
\newblock Margin-based parallel corpus mining with multilingual sentence
  embeddings.
\newblock In {\em Proceedings of the 57th Annual Meeting of the Association for
  Computational Linguistics}, pages 3197--3203, Florence, Italy, July.
  Association for Computational Linguistics.

\bibitem[\protect\citename{Artetxe and Schwenk}2019b]{artetxeSchwenk:2019}
Artetxe, M. and Schwenk, H.
\newblock (2019b).
\newblock Massively multilingual sentence embeddings for zero-shot
  cross-lingual transfer and beyond.
\newblock volume~7, pages 597--610. MIT Press, September.

\bibitem[\protect\citename{Bamman and Smith}2014]{bammanSmith:2014}
Bamman, D. and Smith, N.~A.
\newblock (2014).
\newblock {Unsupervised Discovery of Biographical Structure from Text}.
\newblock {\em Transactions of the Association for Computational Linguistics},
  2:363--376.

\bibitem[\protect\citename{{Barr\'on-Cede{\~n}o} \bgroup et al.\egroup
  }2015]{BarronEtAl:2015}
{Barr\'on-Cede{\~n}o}, A., {Espa{\~n}a-Bonet}, C., {Boldoba}, J., and
  {M\`arquez}, L.
\newblock (2015).
\newblock {A Factory of Comparable Corpora from Wikipedia}.
\newblock In {\em {Proceedings of the 8th Workshop on Building and Using
  Comparable Corpora (BUCC)}}, pages 3--13, Beijing, China, July.

\bibitem[\protect\citename{Basta \bgroup et al.\egroup
  }2019]{basta-etal-2019-evaluating}
Basta, C., Costa-juss{\`a}, M.~R., and Casas, N.
\newblock (2019).
\newblock Evaluating the underlying gender bias in contextualized word
  embeddings.
\newblock In {\em Proceedings of the First Workshop on Gender Bias in Natural
  Language Processing}, pages 33--39, Florence, Italy, August. Association for
  Computational Linguistics.

\bibitem[\protect\citename{Cettolo \bgroup et al.\egroup
  }2012]{cettoloEtAl:2012}
Cettolo, M., Girardi, C., and Federico, M.
\newblock (2012).
\newblock {WIT$^3$: Web Inventory of Transcribed and Translated Talks}.
\newblock In {\em Proceedings of the 16$^{th}$ Conference of the European
  Association for Machine Translation (EAMT)}, pages 261--268, Trento, Italy,
  May.

\bibitem[\protect\citename{Costa-juss\`a}2019]{costajussa:2019}
Costa-juss\`a, M.~R.
\newblock (2019).
\newblock An analysis of gender bias studies in natural language processing.
\newblock {\em Nature Machine Intelligence}, 1.

\bibitem[\protect\citename{\c{S}tef\u{a}nescu \bgroup et al.\egroup
  }2012]{Stefanescuetal:12}
\c{S}tef\u{a}nescu, D., Ion, R., and Hunsicker, S.
\newblock (2012).
\newblock {Hybrid Parallel Sentence Mining from Comparable Corpora}.
\newblock In {\em {Proceedings of the 16th Annual Conference of the European
  Association for Machine Translation (EAMT 2012)}}, Trento, Italy. {European
  Association for Machine Translation }.

\bibitem[\protect\citename{Fleiss and others}1971]{fleiss1971mns}
Fleiss, J. et~al.
\newblock (1971).
\newblock {Measuring nominal scale agreement among many raters}.
\newblock {\em Psychological Bulletin}, 76(5):378--382.

\bibitem[\protect\citename{Font and Costa{-}juss{\`{a}}}2019]{font:2019}
Font, J.~E. and Costa{-}juss{\`{a}}, M.~R.
\newblock (2019).
\newblock Equalizing gender biases in neural machine translation with word
  embeddings techniques.
\newblock {\em CoRR}, abs/1901.03116.

\bibitem[\protect\citename{Graells-Garrido \bgroup et al.\egroup
  }2015]{Graells-GarridoEtAl:2015}
Graells-Garrido, E., Lalmas, M., and Menczer, F.
\newblock (2015).
\newblock {First Women, Second Sex: Gender Bias in Wikipedia}.
\newblock In {\em Proceedings of the 26th ACM Conference on Hypertext \&\#38;
  Social Media}, HT '15, pages 165--174, New York, NY, USA. ACM.

\bibitem[\protect\citename{Habash \bgroup et al.\egroup }2017]{arabAcquis:2017}
Habash, N., Zalmout, N., Taji, D., Hoang, H., and Alzate, M.
\newblock (2017).
\newblock {A Parallel Corpus for Evaluating Machine Translation between Arabic
  and European Languages}.
\newblock In {\em Proceedings of the 15th Conference of the European Chapter of
  the Association for Computational Linguistics: Volume 2, Short Papers}, pages
  235--241, Valencia, Spain, April. Association for Computational Linguistics.

\bibitem[\protect\citename{Koehn \bgroup et al.\egroup }2009]{koehnEtAl:2009}
Koehn, P., Birch, A., and Steinberger, R.
\newblock (2009).
\newblock {462 Machine Translation Systems for Europe}.
\newblock In {\em Proceedings of the Twelfth Machine Translation Summit}, pages
  65--72. Association for Machine Translation in the Americas, AMTA.

\bibitem[\protect\citename{Koehn \bgroup et al.\egroup
  }2019]{koehn-etal-2019-findings}
Koehn, P., Guzm{\'a}n, F., Chaudhary, V., and Pino, J.
\newblock (2019).
\newblock Findings of the {WMT} 2019 shared task on parallel corpus filtering
  for low-resource conditions.
\newblock In {\em Proceedings of the Fourth Conference on Machine Translation
  (Volume 3: Shared Task Papers, Day 2)}, pages 54--72, Florence, Italy,
  August. Association for Computational Linguistics.

\bibitem[\protect\citename{L{\"{a}}ubli \bgroup et al.\egroup
  }2018]{lubliEtAl:2018}
L{\"{a}}ubli, S., Sennrich, R., and Volk, M.
\newblock (2018).
\newblock Has machine translation achieved human parity? {A} case for
  document-level evaluation.
\newblock In Ellen Riloff, et~al., editors, {\em Proceedings of the 2018
  Conference on Empirical Methods in Natural Language Processing, Brussels,
  Belgium, October 31 - November 4, 2018}, pages 4791--4796. Association for
  Computational Linguistics.

\bibitem[\protect\citename{{Park} \bgroup et al.\egroup }2018]{parkEtAl:2018}
{Park}, J.~H., {Shin}, J., and {Fung}, P.
\newblock (2018).
\newblock {Reducing Gender Bias in Abusive Language Detection}.
\newblock {\em ArXiv e-prints}, August.

\bibitem[\protect\citename{Plamada and Volk}2012]{Plamada:12}
Plamada, M. and Volk, M.
\newblock (2012).
\newblock {Towards a Wikipedia-extracted alpine corpus}.
\newblock In {\em Proceedings of the 5th Workshop on Building and Using
  Comparable Corpora: Language Resources for Machine Translation in
  Less-Resourced Languages and Domains}, pages 81--87, Istanbul, Turkey, May.

\bibitem[\protect\citename{Pouliquen \bgroup et al.\egroup
  }2003]{pouliquen:2003}
Pouliquen, B., Steinberger, R., and Ignat, C.
\newblock (2003).
\newblock {Automatic Identification of Document Translations in Large
  Multilingual Document Collections}.
\newblock In {\em Proceedings of the International Conference on Recent
  Advances in Natural Language Processing (RANLP-2003)}, pages 401--408,
  Borovets, Bulgaria.

\bibitem[\protect\citename{{Prates} \bgroup et al.\egroup
  }2018]{pratesEtAl:2018}
{Prates}, M.~O.~R., {Avelar}, P.~H.~C., and {Lamb}, L.
\newblock (2018).
\newblock {Assessing Gender Bias in Machine Translation -- A Case Study with
  Google Translate}.
\newblock {\em ArXiv e-prints}, September.

\bibitem[\protect\citename{Reagle and Rhue}2011]{reagleRhue:2011}
Reagle, J. and Rhue, L.
\newblock (2011).
\newblock {Gender Bias in Wikipedia and Britannica}.
\newblock {\em {International Journal of Communication}}, 5(0).

\bibitem[\protect\citename{Rudinger \bgroup et al.\egroup
  }2018]{rudingerEtAl:2018}
Rudinger, R., Naradowsky, J., Leonard, B., and {Van Durme}, B.
\newblock (2018).
\newblock {Gender Bias in Coreference Resolution}.
\newblock In {\em Proceedings of the Annual Meeting of the North American
  Association of Computational Linguistics (NAACL)}.

\bibitem[\protect\citename{Schwenk \bgroup et al.\egroup
  }2019]{DBLP:journals/corr/abs-1907-05791}
Schwenk, H., Chaudhary, V., Sun, S., Gong, H., and Guzm{\'{a}}n, F.
\newblock (2019).
\newblock Wikimatrix: Mining 135m parallel sentences in 1620 language pairs
  from wikipedia.
\newblock {\em CoRR}, abs/1907.05791.

\bibitem[\protect\citename{Skadi\c{n}a \bgroup et al.\egroup
  }2012]{Skadinaetal:12}
Skadi\c{n}a, I., Aker, A., Mastropavlos, N., Su, F., Tufiș, D., Verlic, M.,
  Vasi\c{l}jevs, A., Babych, B., Clough, P., Gaizauskas, R., Glaros, N.,
  Paramita, M.~L., and Pinnis, M.
\newblock (2012).
\newblock Collecting and using comparable corpora for statistical machine
  translation.
\newblock In Nicoletta Calzolari, et~al., editors, {\em {Proceedings of the
  Eighth International Conference on Language Resources and Evaluation (LREC
  2012)}}, Istanbul, Turkey, May. {European Language Resources Association
  (ELRA)}.

\bibitem[\protect\citename{Stanovsky \bgroup et al.\egroup
  }2019]{stanovsky-etal-2019-evaluating}
Stanovsky, G., Smith, N.~A., and Zettlemoyer, L.
\newblock (2019).
\newblock Evaluating gender bias in machine translation.
\newblock In {\em Proceedings of the 57th Annual Meeting of the Association for
  Computational Linguistics}, pages 1679--1684, Florence, Italy, July.
  Association for Computational Linguistics.

\bibitem[\protect\citename{Wagner \bgroup et al.\egroup }2016]{wagnerEtAl:2016}
Wagner, C., Graells-Garrido, E., Garcia, D., and Menczer, F.
\newblock (2016).
\newblock Women through the glass ceiling: gender asymmetries in wikipedia.
\newblock {\em EPJ Data Science}, 5(1):5, Mar.

\bibitem[\protect\citename{{Webster} \bgroup et al.\egroup
  }2018]{websterEtAl:2018}
{Webster}, K., {Recasens}, M., {Axelrod}, V., and {Baldridge}, J.
\newblock (2018).
\newblock {Mind the GAP: A Balanced Corpus of Gendered Ambiguous Pronouns}.
\newblock {\em ArXiv e-prints}, October.

\bibitem[\protect\citename{Yasuda and Sumita}2008]{Yasuda:08}
Yasuda, K. and Sumita, E.
\newblock (2008).
\newblock {Method for Building {Sentence-Aligned} Corpus from Wikipedia}.
\newblock In {\em {Proceedings of the AAAI Workshop on Wikipedia and Artificial
  Intelligence: An Evolving Synergy}}, pages 64--66, Menlo Park, CA.

\bibitem[\protect\citename{Zhao \bgroup et al.\egroup }2018]{zhaoEtAl:2018}
Zhao, J., Wang, T., Yatskar, M., Ordonez, V., and Chang, K.-W.
\newblock (2018).
\newblock Gender bias in coreference resolution: Evaluation and debiasing
  methods.
\newblock In {\em Proceedings of the 2018 Conference of the North American
  Chapter of the Association for Computational Linguistics: Human Language
  Technologies, Volume 2 (Short Papers)}, pages 15--20. Association for
  Computational Linguistics.

\end{thebibliography}


\end{document}